\documentclass[10pt]{article}

\usepackage[a4paper,margin=1in]{geometry}
\usepackage{amsmath,amssymb,amsthm}
\usepackage{microtype}
\usepackage{graphicx}
\usepackage{booktabs}
\usepackage{enumitem}
\usepackage{xcolor}
\usepackage[hidelinks,colorlinks=true,linkcolor=black,citecolor=blue,urlcolor=blue]{hyperref}
\usepackage{titlesec}
\usepackage{tikz}
\usetikzlibrary{positioning,arrows.meta}

\titleformat{\section}{\normalfont\large\bfseries}{\thesection}{0.6em}{}
\titleformat{\subsection}{\normalfont\bfseries}{\thesubsection}{0.6em}{}
\newcommand{\Rb}{\mathbb{R}}
\newcommand{\T}{^{\!\top}}
\newcommand{\Peff}{P_{\mathrm{eff}}}
\newcommand{\Qeff}{Q_{\mathrm{eff}}}
\newcommand{\Pbasis}{P_{\mathrm{basis}}}
\newcommand{\Qbasis}{Q_{\mathrm{basis}}}
\newcommand{\Pbases}{P_{\mathrm{bases}}}

\title{\textbf{Ortho-Hydra: Orthogonalized Experts for DiT LoRA}\\[2pt]
\large A structural fix for the MoE-LoRA cold-start deadlock}
\author{Seunghyun Ji\\\small\texttt{standingbehindnv@gmail.com}}
\date{April 2026}

\begin{document}
\maketitle

\begin{abstract}
LoRA fine-tuning of diffusion transformers (DiT) on multi-style data
suffers from \emph{style bleed}: a single low-rank residual cannot
represent several distinct artist fingerprints, and the optimiser
converges to their average. Mixture-of-experts LoRA in the HydraLoRA
style replaces the up-projection with $E$ heads under a router, but
when every expert is zero-initialized the router receives identical
gradient from each head and remains at the uniform prior. The
experts then evolve permutation-symmetrically, and the network trains
as a single rank-$r$ LoRA at $E{\times}$ the cost. We present
\textbf{Ortho-Hydra}, a re-parameterization that combines an OFT-style
Cayley-orthogonal shared basis with per-expert \emph{disjoint output
subspaces} carved from the top-$(Er)$ left singular vectors of the
pretrained weight. Disjointness makes the router's per-expert score
non-degenerate at step~$0$, so specialization receives gradient signal
before any expert has trained. We test the predicted
deadlock on a DiT pipeline by comparing two HydraLoRA baselines, a
zero-initialized shared-basis variant and the original
$\sigma{=}0.1$ Gaussian-jitter mitigation, against Ortho-Hydra under a
matched optimiser, dataset, and step budget. Neither baseline leaves
the uniform prior within the first $1\text{k}$ steps; Ortho-Hydra
begins de-uniformising within the first few hundred. End-task generation
quality on multi-style data is out of scope; we report the construction,
the cold-start mechanism, and the routing dynamics it changes.
Code: \url{https://github.com/sorryhyun/anima_lora}.
\end{abstract}

\section{Introduction}

Low-rank adaptation~\cite{hu2022lora} replaces a frozen weight $W_0
\in \Rb^{d_{\mathrm{out}}\times d_{\mathrm{in}}}$ with a residual
$y = W_0 x + s\,B A x$, where $A \in \Rb^{r\times d_{\mathrm{in}}}$ and
$B \in \Rb^{d_{\mathrm{out}}\times r}$ are trained and $s = \alpha/r$
is a fixed scalar. For diffusion-transformer fine-tuning at $r \in [4,
64]$ this is parameter- and storage-efficient, and it
has become the default adapter format for community model distribution.

Two well-known weaknesses of LoRA compound on \emph{multi-style}
training data. First, the residual $BA$ can drift in directions that
fight the pretrained weight, so style adaptation also degrades anatomy
or text rendering. Second, with a single shared $(A,B)$ pair the
optimiser's best rank-$r$ compromise across $K$ artists is the
subspace that captures \emph{common} structure, so the union becomes
a blended average rather than any individual fingerprint. The first
weakness motivates orthogonal fine-tuning~\cite{qiu2023oft,liu2024boft,
wu2026psoft}, which constrains the delta to a structurally orthogonal
subspace; the second motivates mixture-of-experts
LoRA~\cite{tian2024hydralora}, which equips the up-projection with
multiple heads and a learnt router so distinct sample clusters can
push different heads in different directions.

The two fixes solve different problems and compose poorly when stacked
naively. Section~\ref{sec:method} describes an MoE-LoRA cold-start
deadlock that arises when a shared orthogonal basis is paired with
multiple zero-initialized expert heads, and a structural fix:
partitioning the top-$(Er)$ singular directions into disjoint
per-expert slices breaks the deadlock at initialization, before the
router has trained.

This is a focused methods report. The construction is implemented and
trains in our DiT pipeline. We test the deadlock prediction against
two shared-basis HydraLoRA baselines under identical optimiser,
dataset, and step budgets (Figure~\ref{fig:router-dynamics}). The
predicted uniform-routing plateau is directly observable in router
entropy, and the original $\sigma$-jitter mitigation does not escape
it on the same budget. End-task generation quality on multi-style data
and sensitivity ablations over $E$, $r$, and the balance-loss weight
are out of scope.

\paragraph{Contributions.}
\begin{enumerate}[topsep=2pt,itemsep=0pt,leftmargin=1.2em]
\item We identify a cold-start deadlock specific to MoE-LoRA with a
shared orthogonal basis: every expert lives in the same rank-$r$
column span, the router's per-expert score is an orthogonal-matrix
inner product that cannot vanish, and gradients to differentiate
experts are degenerate at initialization.
\item We propose \emph{disjoint SVD-slice expert bases}: the
top-$(Er)$ left singular vectors of $W_0$ are partitioned into $E$
orthonormal slices, one per expert, and rotated only \emph{within}
their slice by a per-expert Cayley matrix. Cross-expert orthogonality
is preserved exactly through training.
\item We pair this with a \emph{layer-local} router that reads the
RMS-pooled rank-$r$ bottleneck activation rather than the raw input
or the text embedding, and explain why this aggregator is the only
one of three obvious choices that survives the long-sequence
DC-cancellation problem in DiT.
\end{enumerate}

\section{Background}
\label{sec:background}

\paragraph{LoRA.} For frozen $W_0$, the residual is $y = W_0 x + s\,
B A x$ with $s = \alpha/r$. Standard practice~\cite{hu2022lora}
initializes $B = 0$ so that step~$0$ reproduces the base model exactly.

\paragraph{Orthogonal fine-tuning (OFT, BOFT, PSOFT).} These methods
replace the free delta $BA$ with a structurally constrained one.
OFT~\cite{qiu2023oft} multiplies $W_0$ by a learned orthogonal matrix;
BOFT~\cite{liu2024boft} factors that matrix as a butterfly product
for parameter efficiency; PSOFT~\cite{wu2026psoft} restricts the
rotation to the \emph{principal subspace} of $W_0$ via a frozen SVD
basis with a Cayley-parameterized rotation inside that subspace.
Cayley parameterization~\cite{lezcano2019cayley} guarantees
orthogonality at every step: for skew-symmetric $A = S - S\T$,
$R = (I-A)(I+A)^{-1}$ satisfies $R\T R = I$ exactly, with no
regularization hyperparameter.

\paragraph{HydraLoRA.} The closest prior work~\cite{tian2024hydralora}
keeps the shared down-projection $A$ but replaces $B$ with $E$
parallel up-heads $\{B_i\}_{i=1}^E$ and a router $g(\cdot) \in
\Delta^{E-1}$ that emits a per-sample softmax gate. The effective
up-projection $B_{\mathrm{eff}}(x) = \sum_i g_i(x)\, B_i$ becomes
sample-dependent. With a Switch-Transformer~\cite{fedus2022switch}
balance loss to keep all experts alive, this admits per-cluster
specialization in principle.

\paragraph{Soft expert-specialization regularizers.} A separate
line of work~\cite{guo2025advancing} attacks expert overlap from the
loss side rather than the architecture: an output-orthogonality term
penalising inner products between selected experts' per-token outputs,
and a routing-variance term encouraging decisive gates, both stacked
on the standard auxiliary balance loss. The motivating regime is a
pretrained MoE whose experts already differ but whose router is being
driven back toward uniformity by the balance objective during
fine-tuning. We compare against this line in
\S\ref{sec:discussion}, after the construction is on the table.

\paragraph{The cold-start deadlock.} HydraLoRA initializes every $B_i$
to zero so the residual is zero at step~$0$. Under a near-uniform
router every expert receives \emph{identical} gradient, and the
experts evolve permutation-symmetrically: they remain identical
indefinitely, the router never receives a differentiating signal, and
the network trains as a single rank-$r$ LoRA averaged over $E$ heads.
The original mitigation is \emph{expert warm-up}: for the first
$\rho \cdot N_{\mathrm{steps}}$ steps, only a randomly chosen expert
per module receives gradient at each step, breaking the symmetry by
schedule. This is fragile to schedule choices and provides no signal
to the router until the $B_i$ have drifted apart.

\section{Method: Ortho-Hydra}
\label{sec:method}

\begin{figure}[t]
\centering
\resizebox{\linewidth}{!}{%
\begin{tikzpicture}[
  font=\footnotesize,
  >={Stealth[length=1.6mm,width=1.2mm]},
  box/.style={draw, rounded corners=1pt, minimum height=4mm,
              minimum width=15mm, fill=gray!6, inner sep=1pt},
  small/.style={draw, rounded corners=1pt, minimum height=4mm,
                minimum width=5mm, fill=gray!6, inner sep=0pt,
                font=\scriptsize},
  ph/.style={minimum height=4mm, minimum width=4mm, inner sep=0pt,
             font=\scriptsize},
  router/.style={draw, rounded corners=1pt, minimum height=4mm,
                 minimum width=12mm, fill=blue!10, inner sep=1pt,
                 font=\scriptsize},
  dot/.style={circle, draw, fill=gray!18, inner sep=0pt,
              minimum size=2mm},
  plus/.style={circle, draw, inner sep=0pt, minimum size=4mm,
               font=\scriptsize},
  io/.style={font=\itshape},
  arr/.style={->, thin},
  thinarr/.style={->, thin, gray!60!black},
  panhead/.style={font=\bfseries\footnotesize},
  lab/.style={font=\scriptsize, gray!45!black},
  node distance=2.4mm,
]

\begin{scope}[local bounding box=panA]
  \node[panhead] (titA) {(a) Plain LoRA};
  \node[io, below=of titA] (xa) {$x$};
  \node[box, below=of xa] (Aa) {$A$};
  \node[dot, below=of Aa] (la) {};
  \node[lab, right=1mm of la] {$\ell\,{\in}\,\mathbb{R}^r$};
  \node[box, below=of la] (Ba) {$B$\ \scriptsize{(init $0$)}};
  \node[plus, below=4mm of Ba] (sa) {\scriptsize$+$};
  \node[box, fill=white, left=5mm of sa] (Wa) {$W_0\,x$};
  \node[io, below=2mm of sa] (ya) {$y$};
  \draw[arr] (xa)--(Aa); \draw[arr] (Aa)--(la);
  \draw[arr] (la)--(Ba); \draw[arr] (Ba)--(sa);
  \draw[arr] (Wa)--(sa); \draw[arr] (sa)--(ya);
  \node[draw, thin, minimum width=3.5mm, minimum height=22mm,
        right=12mm of la, anchor=north] (cspA) {};
  \node[lab, above=0.4mm of cspA] {$d_{\mathrm{out}}$};
  \fill[red!55] ([yshift=-9mm]cspA.north west)
                rectangle ([yshift=-13mm]cspA.north east);
  \node[lab, right=0.6mm of cspA, anchor=west] {$\mathrm{col}(B)$};
\end{scope}

\begin{scope}[local bounding box=panB, shift={(58mm,0)}]
  \node[panhead] (titB) {(b) HydraLoRA};
  \node[io, below=of titB] (xb) {$x$};
  \node[box, below=of xb] (Ab) {$A$ \scriptsize{(shared)}};
  \node[dot, below=of Ab] (lb) {};
  \node[lab, right=1mm of lb] {$\ell$};
  \node[small, below=4mm of lb, xshift=-9mm] (B1) {$B_1$};
  \node[small, right=0.8mm of B1] (B2) {$B_2$};
  \node[ph, right=0.8mm of B2] (Bd) {$\cdots$};
  \node[small, right=0.8mm of Bd] (BE) {$B_E$};
  \node[plus, below=14mm of lb] (sb) {\scriptsize$+$};
  \node[box, fill=white, left=5mm of sb] (Wb) {$W_0\,x$};
  \node[io, below=2mm of sb] (yb) {$y$};
  \node[router, left=7mm of B1] (rb) {router\,$g(x)$};
  \draw[arr] (xb)--(Ab); \draw[arr] (Ab)--(lb);
  \draw[arr] (lb)--(B1.north);
  \draw[arr] (lb)--(B2.north);
  \draw[arr] (lb)--(BE.north);
  \draw[thinarr] (B1.south)--(sb);
  \draw[thinarr] (B2.south)--(sb);
  \draw[thinarr] (BE.south)--(sb);
  \draw[arr] (Wb)--(sb); \draw[arr] (sb)--(yb);
  \draw[thinarr] (xb.west) to[out=180,in=90] (rb.north);
  \draw[thinarr] (rb.east) -- node[lab,above=-0.2mm]{$g$} (B1.west);
  \node[draw, thin, minimum width=3.5mm, minimum height=22mm,
        right=12mm of lb, anchor=north] (cspB) {};
  \node[lab, above=0.4mm of cspB] {$d_{\mathrm{out}}$};
  \fill[red!55, opacity=0.65]
        ([yshift=-9mm]cspB.north west)
        rectangle ([yshift=-13mm]cspB.north east);
  \fill[blue!55, opacity=0.55]
        ([yshift=-9.4mm]cspB.north west)
        rectangle ([yshift=-12.6mm]cspB.north east);
  \fill[green!60!black, opacity=0.45]
        ([yshift=-9.8mm]cspB.north west)
        rectangle ([yshift=-12.2mm]cspB.north east);
  \node[lab, below=0.4mm of cspB] {$\mathrm{col}(B_e)$};
\end{scope}

\begin{scope}[local bounding box=panC, shift={(118mm,0)}]
  \node[panhead] (titC) {(c) Ortho-Hydra};
  \node[io, below=of titC] (xc) {$x$};
  \node[box, below=of xc] (Qc) {$R_q\, Q_{\mathrm{basis}}$};
  \node[dot, below=of Qc] (lc) {};
  \node[lab, right=1mm of lc] {$\ell$};
  \node[router, below=2mm of Qc, xshift=-15mm, anchor=east]
        (rc) {router\,$g(\bar\ell)$};
  \node[box, below=of lc] (Lam) {$\odot\,\mathrm{diag}(\lambda)$};
  \node[small, below=4mm of Lam, xshift=-9mm, fill=red!22] (P1) {$P_1$};
  \node[small, right=0.8mm of P1, fill=blue!22] (P2) {$P_2$};
  \node[ph, right=0.8mm of P2] (Pd) {$\cdots$};
  \node[small, right=0.8mm of Pd, fill=green!28!white] (PE) {$P_E$};
  \node[plus, below=22mm of lc] (sc) {\scriptsize$+$};
  \node[box, fill=white, left=5mm of sc] (Wc) {$W_0\,x$};
  \node[io, below=2mm of sc] (yc) {$y$};
  \draw[arr] (xc)--(Qc); \draw[arr] (Qc)--(lc);
  \draw[arr] (lc)--(Lam);
  \draw[arr] (Lam)--(P1.north);
  \draw[arr] (Lam)--(P2.north);
  \draw[arr] (Lam)--(PE.north);
  \draw[thinarr] (P1.south)--(sc);
  \draw[thinarr] (P2.south)--(sc);
  \draw[thinarr] (PE.south)--(sc);
  \draw[arr] (Wc)--(sc); \draw[arr] (sc)--(yc);
  \draw[thinarr] (lc.west) to[out=180,in=0]
        node[lab,pos=0.5,above=-0.4mm]{rmspool} (rc.east);
  \draw[thinarr] (rc.south) to[out=270,in=180]
        node[lab,pos=0.85,above=-0.4mm]{$g$} (P1.west);
  \node[draw, thin, minimum width=3.5mm, minimum height=22mm,
        right=12mm of lc, anchor=north] (cspC) {};
  \node[lab, above=0.4mm of cspC] {$d_{\mathrm{out}}$};
  \fill[red!60]
        ([yshift=-3mm]cspC.north west)
        rectangle ([yshift=-7mm]cspC.north east);
  \fill[blue!60]
        ([yshift=-7.5mm]cspC.north west)
        rectangle ([yshift=-11mm]cspC.north east);
  \fill[gray!40]
        ([yshift=-11.5mm]cspC.north west)
        rectangle ([yshift=-15mm]cspC.north east);
  \fill[green!60!black]
        ([yshift=-15.5mm]cspC.north west)
        rectangle ([yshift=-19mm]cspC.north east);
  \node[lab, below=0.4mm of cspC] {$\mathrm{col}(P_e)$};
\end{scope}

\end{tikzpicture}}
\caption{Architecture comparison.
\textbf{(a) Plain LoRA:} a single $(A, B)$ pair; $B$ writes the
residual into one $r$-dimensional band of
$\mathbb{R}^{d_{\mathrm{out}}}$ (right cartoon).
\textbf{(b) HydraLoRA:} shared $A$ with $E$ parallel up-heads
$\{B_e\}$ mixed by a router $g(x)$ under a Switch-Transformer balance
loss. With $B_e\!=\!0$ at initialization, every head shares the same
column span (overlapping bands at right) and the per-expert router
score is degenerate; specialization receives no gradient signal until
the symmetry is broken externally.
\textbf{(c) Ortho-Hydra:} the down-projection is a Cayley-rotated SVD
basis $R_q\,\Qbasis$; the up-projection is a sum
$\sum_e g_e\, \Peff[e]$ over $E$ \emph{disjoint} orthonormal slices
of the top-$(Er)$ left singular vectors of $W_0$, each rotated only
within its own slice by a per-expert Cayley matrix. The
$\Pbases[i]\T\Pbases[j]\!=\!\mathbf{0}$ invariant
(non-overlapping bands at right) is preserved exactly through training,
and the layer-local router reads the RMS-pooled rank-$r$ activation
$\bar\ell$ rather than the raw input.}
\label{fig:archs}
\end{figure}
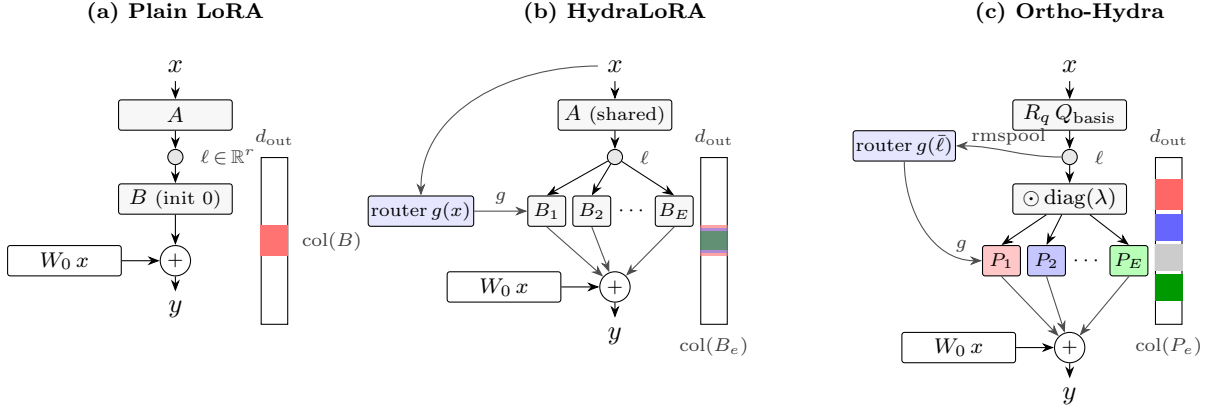

\subsection{Architecture}

For each adapted Linear $W_0 \in \Rb^{d_{\mathrm{out}}\times
d_{\mathrm{in}}}$ we store the components in
Table~\ref{tab:components}; Figure~\ref{fig:archs} contrasts the
resulting architecture against plain LoRA and HydraLoRA.

\begin{center}
\small
\begin{tabular}{@{}lll@{}}
\toprule
Component & Shape & Trainable? \\
\midrule
$\Qbasis$ & $(r,\, d_{\mathrm{in}})$ & frozen buffer (shared) \\
$\Pbases$ & $(E,\, d_{\mathrm{out}},\, r)$ & frozen buffer (per-expert disjoint) \\
$S_q$ & $(r,\, r)$ & yes, shared \\
$S_p$ & $(E,\, r,\, r)$ & yes, per-expert \\
$\lambda$ & $(1,\, r)$ & yes, shared \\
$\theta_{\mathrm{router}}$ & $\Rb^{(r+d_\sigma)\times E} \times \Rb^E$ & yes, layer-local \\
\bottomrule
\end{tabular}
\end{center}
\refstepcounter{table}\label{tab:components}

Compute the thin SVD of $W_0$ via randomised
SVD~\cite{halko2011randomized}, $W_0 \approx U\Sigma V\T$, with $U
\in \Rb^{d_{\mathrm{out}}\times q}$ and $V \in
\Rb^{d_{\mathrm{in}}\times q}$ at $q = Er + 6$. Set
\[
\Qbasis \leftarrow V_{:,1:r}\T,
\qquad
\Pbases[e] \leftarrow U_{:,\,(e-1)r+1:\,er}
\;\;\text{for}\;\; e=1,\dots,E.
\]
$\Qbasis$ has orthonormal rows, and the $\Pbases[e]$ are mutually
orthonormal slices of $U$: $\Pbases[i]\T \Pbases[j] = \mathbf{0}$
for $i \neq j$ and $= I_r$ for $i=j$. This invariant underlies the
deadlock-breaking argument of \S\ref{sec:deadlock-fix}.

The trainable rotations are skew-symmetric seeds. Given $S_q$ and
$S_p[e]$, define $A_q = S_q - S_q\T$ and $A_p[e] = S_p[e] - S_p[e]\T$,
and form Cayley orthogonal matrices
\[
R_q = (I - A_q)(I + A_q)^{-1},
\qquad
R_p[e] = (I - A_p[e])(I + A_p[e])^{-1}.
\]
We compute the inverse with \texttt{torch.linalg.solve} rather than a
Neumann truncation~\cite{wu2026psoft}: at $r \leq 64$ the cost is
negligible and the solve is unconditionally correct, whereas a
$K$-term Neumann series silently diverges when $\lVert A\rVert > 1$
and breaks orthogonality without a loss-side signal.

The effective bases $\Qeff = R_q \Qbasis$ (rows orthonormal, shared)
and $\Peff[e] = \Pbases[e]\, R_p[e]$ (columns orthonormal within
slice $e$) preserve cross-expert orthogonality exactly through
training: $\Peff[i]\T \Peff[j] = R_p[i]\T\, \Pbases[i]\T\,
\Pbases[j]\, R_p[j] = \mathbf{0}$ for $i \neq j$, and $= R_p[i]\T
R_p[j]$ otherwise.

\subsection{Forward and routing}

For input $x \in \Rb^{B\times L\times d_{\mathrm{in}}}$,
\begin{align*}
\ell &= x\,\Qeff\T \in \Rb^{B\times L\times r}, \\
g    &= \mathrm{softmax}\bigl(\theta_{\mathrm{router}}\bigl[\mathrm{rmspool}_L(\ell)\,\big\Vert\,\phi(\sigma)\bigr]\bigr) \in \Delta^{E-1}, \\
\ell &\leftarrow \ell\,\odot\,\lambda, \\
P_{\mathrm{combined}}(b) &= \sum_{e=1}^{E} g_{b,e}\,\Peff[e] \quad\in\Rb^{d_{\mathrm{out}}\times r}, \\
y    &= W_0 x + s\,P_{\mathrm{combined}}\,\ell.
\end{align*}
Here $\mathrm{rmspool}_L(\ell)_j = \sqrt{(1/L)\sum_\ell
\ell_{\cdot,\ell,j}^2}$ is RMS pool over the sequence dimension, and
$\phi(\sigma)$ are sinusoidal features of the diffusion noise level
(optional, with zero-init columns in the router so step-$0$ behaviour
matches the no-$\sigma$ case).

\paragraph{Why RMS over the rank-$r$ bottleneck.}
Pooling over a $\sim$$4096$-token DiT sequence breaks zero-mean
features under mean-pooling: per-channel std after the mean is
$\sigma_x / \sqrt{L} \approx 0.008$, so the pooled vector is
near-identical across samples and the router sees effectively
constant input. Pooling \emph{after} the rank-$r$ bottleneck rather
than over raw $d_{\mathrm{in}}$-wide activations also matters,
because DiT layer inputs have heavy DC-bias outlier channels (peak/mean
ratios in the $80$--$96\times$ range) that saturate softmax in bf16
under max pool, while the rank-$r$ space is bounded by
$\lVert\Qbasis\rVert\cdot\lVert x\rVert$ with no such outliers. RMS,
unlike mean, does not cancel zero-mean signals by $\sqrt{L}$ because
signs square to positive before averaging. Of the three pool/space
combinations we tried, this is the only one whose router weight norm
moves from initialization; the other two leave the router
indistinguishable from random init at end of training.

\subsection{Why disjoint slices break the deadlock}
\label{sec:deadlock-fix}

The router's per-expert score before softmax can be written, ignoring
the $\sigma$-feature columns, as
\[
\mathrm{score}_e \;=\; w_e\T \,\mathrm{rmspool}_L\bigl(x\,\Qeff\T\bigr)
\]
with $w_e \in \Rb^r$ a slice of the router weight. At init $\lambda
= 0$, so the residual is zero and the gradient into $w_e$ comes from
the chain rule through $\Peff[e]$ once $\lambda$ takes its first
non-zero step. The relative magnitudes of these gradients are
governed, to first order, by the Gram matrix $\Peff[i]\T \Peff[j]$.

\begin{itemize}[topsep=2pt,itemsep=0pt,leftmargin=1.2em]
\item \emph{Shared basis.} If every expert uses the same $\Pbasis$,
then $\Peff[i]\T \Peff[j] = R_p[i]\T R_p[j]$ is an $r\times r$
orthogonal matrix and cannot be zero; at zero-init $S_p$ it is the
identity. Every expert produces the same per-token output direction;
gradients into the $w_e$ are near-identical; softmax gates stay near
$1/E$; the gradient into $S_p[e]$ is the same for every $e$; experts
evolve permutation-symmetrically. This is the deadlock.

\item \emph{Disjoint slices.} If $\Pbases[i]\T \Pbases[j] =
\mathbf{0}$ for $i \neq j$, then so is $\Peff[i]\T \Peff[j]$ for
every value of $S_p$. Each expert writes into a distinct output
subspace; the per-expert components of the upstream gradient are
orthogonal in $\Rb^{d_{\mathrm{out}}}$, so their projections onto
each $w_e$ are genuinely different. The router has signal before any
expert has trained.
\end{itemize}

The construction is structural rather than scheduled, and operates
one level below loss-side specialization
regularizers~\cite{guo2025advancing}: we revisit this comparison in
\S\ref{sec:discussion}. It does not replace the Switch-Transformer
balance loss, which still keeps all experts alive against the
tendency of a sharp router to collapse onto one; it removes the
bootstrapping problem that the balance loss alone cannot solve.

\paragraph{When disjointness is impossible.}
The construction requires $\min(d_{\mathrm{out}}, d_{\mathrm{in}})
\geq Er$. For DiT-scale attention and MLP weights with $d \in \{1024,
3072, 4096\}$ and our default $E=4$, this is satisfied at every rank
we have used ($r \leq 64$). On narrow projections that violate the
inequality, the implementation falls back to a shared-basis $\Pbases$
replicated across experts (with a runtime warning); in that fallback
the only symmetry-breaker is the original expert warm-up schedule,
and running with warm-up disabled is unsafe.

\section{Experiments}
\label{sec:experiments}

\begin{figure}[t]
\centering
\includegraphics[width=\linewidth]{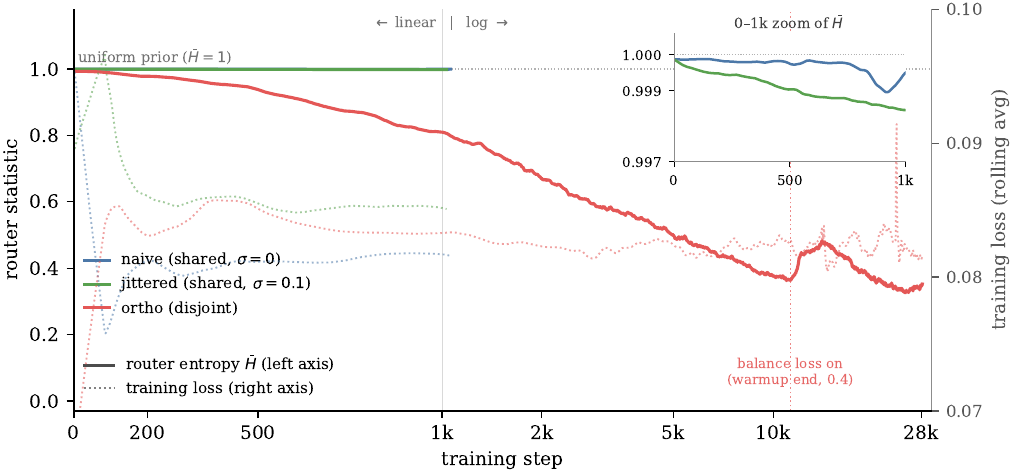}
\caption{Cold-start router dynamics across three HydraLoRA variants
($E\!=\!12$, balance-loss weight $5\!\times\!10^{-7}$ with warmup
ratio $0.4$, AdamW, identical dataset and optimiser schedule). Solid:
mean normalised router entropy $\bar H$ (left axis); dotted: training
loss (right axis), zoomed to $[0.07, 0.10]$ as a proof-of-life
signal. \emph{Naive} (shared basis, zero-init experts) and
\emph{jittered} (shared basis with $\sigma\!=\!0.1$ Gaussian on
$B_i$, the original HydraLoRA mitigation~\cite{tian2024hydralora})
remain pinned at $\bar H\!\approx\!1$ for the full $1\text{k}$ steps;
\emph{ortho} (disjoint slices) leaves the uniform prior by step
$\sim\!200$ and converges to $\bar H\!\approx\!0.35$ over $28\text{k}$
steps. Inset: $0$--$1\text{k}$ window magnified to
$\bar H\!\in\![0.997, 1.0006]$, showing that jitter buys only modest
extra drift over naive. The vertical at step $11266$ marks the
balance-loss weight turning on ($0.4\!\cdot\!28164$ max steps); the
small entropy bump in ortho around that step is the load-balance
penalty pulling a specialized router back toward uniform. Training
loss confirms all three variants reach the same $\sim\!0.08$ band, so
the gap is in routing rather than optimisation. The ortho run used
$r\!=\!48$ versus $r\!=\!32$ for the baselines; the $\bar H$ gap is
too large for a $1.5\!\times$ rank difference to explain, and more
rank does not help shared-basis routing. The naive and jittered runs
were stopped at $1\text{k}$ steps once the plateau at
$\bar H\!\approx\!1$ was unambiguous.}
\label{fig:router-dynamics}
\end{figure}

\paragraph{Base model.}
We adapt \emph{anima-preview3-base}~\cite{anima2026}, a $28$-block,
$2048$-channel DiT ($16$ heads of dim $128$ each, MLP ratio $4$)
trained with a flow-matching velocity-field objective on multi-style
anime data. Text conditioning is a frozen Qwen3-0.6B base encoder,
and the latent codec is the qwen-image VAE. All three modules are
held frozen through training; only the LoRA / MoE adapter is updated.
The bf16 DiT checkpoint is ${\sim}\,4.2$\,GB.

\paragraph{Data.}
The training set contains ${\sim}\,2.4\text{k}$ images covering
several artist styles, each with a free-form caption. We use masked
loss with masks merged from a SAM3 pass and a manga-text-detector pass
(text bubbles excluded from the loss). Latents are pre-cached via the
VAE and text outputs via the Qwen3 encoder, and the DiT is loaded
only after caching to free VRAM. Constant-token bucketing rounds each
sample to ${\sim}\,4096$ latent tokens with zero-padding, giving
\texttt{torch.compile} a single static shape across aspect ratios.

\paragraph{Optimiser and schedule.}
AdamW (fused) at $\mathrm{lr}\!=\!2\!\times\!10^{-5}$, constant
schedule, no warmup. bf16 mixed precision; flow shift $1.0$;
sigmoid timestep sampling; caption dropout $0.1$; T-LoRA timestep
masking enabled on every variant. Batch size $1$ per step at
constant-token-budget $4096$. The full ortho run is $28164$ steps
($12$ epochs over the dataset); the shared-basis baselines are stopped
at $1\text{k}$ steps once the plateau at $\bar H\!\approx\!1$ is
unambiguous.

\paragraph{Adapter configuration.}
HydraLoRA is applied to the two MLP linears
(\texttt{layer1}, \texttt{layer2}) of every DiT block, for
$28\!\times\!2 = 56$ adapted Linears in total, at $E\!=\!12$ experts
and $\alpha\!=\!32$. The router is layer-local with $4$-bucket sigma
features (a sinusoidal encoding of the diffusion noise level fed
alongside the RMS-pooled rank-$r$ activation), and its learning rate
is scaled $10\!\times$ above the adapter LR. The
Switch-Transformer~\cite{fedus2022switch} balance loss has weight
$5\!\times\!10^{-7}$ with warmup ratio $0.4$ (turn-on at step $11266
= 0.4\!\cdot\!28164$) and a per-bucket coefficient of $0.3$ to keep
each sigma-bucket's expert distribution near-uniform. Variant-specific
knobs:
\begin{itemize}[topsep=2pt,itemsep=0pt,leftmargin=1.2em]
\item \emph{naive}: shared-basis HydraLoRA, $r\!=\!32$, $B_i\!=\!0$,
no symmetry-breaker.
\item \emph{jittered}: shared-basis HydraLoRA, $r\!=\!32$,
$B_i\!\sim\!\mathcal{N}(0, 0.1^2)$, the original HydraLoRA
mitigation~\cite{tian2024hydralora}.
\item \emph{ortho}: this construction with disjoint SVD-slice expert
bases, $r\!=\!48$. The rank mismatch with the baselines is a
wall-clock tradeoff; the cold-start signal saturates by $\sim$$200$
steps, so a rank-matched repeat is unlikely to alter the routing
conclusion. We discuss the caveat with the figure.
\end{itemize}

\paragraph{Metrics.}
\emph{Mean normalised router entropy} $\bar H \in [0,\,1]$: per-Linear
softmax entropy divided by $\log E$ and averaged across the $56$
routed Linears, so $\bar H\!=\!1$ at the uniform prior and
$\bar H\!=\!0$ at one-hot routing. \emph{Training loss}: the standard
flow-matching velocity MSE (masked, as configured for training).

\paragraph{Results.}
Figure~\ref{fig:router-dynamics} shows the cold-start router dynamics
across the three variants. The two shared-basis runs remain pinned at
the uniform prior $\bar H\!\approx\!1$ for the full $1\text{k}$ steps
($\bar H$ moves by $<\!0.005$), while
ortho has already left $\bar H\!\approx\!0.81$ by step $1\text{k}$
and converges to $\bar H\!\approx\!0.35$ over $28\text{k}$ steps with
all twelve experts alive at the end. Training loss is essentially
identical across variants ($\sim\!0.08$ end-of-run band, dotted right
axis), so the difference is confined to routing. The de-uniformisation
in ortho is gradual over the first ${\sim}\,2\text{k}$ steps rather
than instantaneous: the router gradient is gated by $\lambda$ leaving
zero, so the deadlock claim is about gradient \emph{direction} (the
per-expert score is non-degenerate from the start) rather than
gradient \emph{magnitude} (initially small, growing with $\lambda$).
The most informative observation is that the jittered baseline barely
outperforms naive: an aggressive $\sigma\!=\!0.1$ Gaussian on $B_i$
that fully randomises the up-projection still does not break the
shared-basis deadlock at the rate disjoint slices do.

\section{Discussion and limitations}
\label{sec:discussion}

\paragraph{Subspace restriction.}
$\Delta W(b)$ is constrained to live in $\mathrm{colspace}(\Pbases)
\subseteq \mathrm{top}\text{-}(Er)$ left singular vectors of $W_0$.
For NLP fine-tuning this is desirable, since the model already knows
the domain and the delta nudges it within the
domain~\cite{wu2026psoft}. For creative fine-tuning of a diffusion
model the assumption is shakier: a new character or art style might
require a $\Delta W$ component outside the principal subspace. Whether
the lost expressiveness matters in the multi-style fine-tuning regime
is an empirical question we do not answer here.

\paragraph{Disjointness is sufficient, not optimal.}
Disjoint slices break the cold-start deadlock, but other choices
(Hadamard, structured random) also satisfy the cross-expert
inner-product condition. The SVD choice is motivated by the same
intuition as PSOFT: the top-$(Er)$ singular directions are where the
pretrained weight already concentrates capacity, so an adapter delta
in those directions is more likely to be useful. This is a hypothesis
rather than a result.

\paragraph{Why $\sigma$-jitter is brittle.}
The Gaussian-noise mitigation has two failure modes that the
disjoint-slice construction sidesteps by having no analogous knob.
First, the noise is loss-agnostic: it perturbs each $B_i$ in random
directions in $\Rb^{d_{\mathrm{out}}}$, but those directions are
generically unaligned with the directions in which the multi-style
data wants to push different experts, and the shared-basis gradient
flattens whatever residual asymmetry the noise created. The
$1\text{k}$-step jittered run in Figure~\ref{fig:router-dynamics}
executes entirely inside the balance-loss warmup region (turn-on at
step $11266 = 0.4\!\cdot\!N$), yet $\bar H$ moves by $<\!0.005$ over
that window: shared-basis dynamics flatten the noise on their own,
with no help from the load balancer. Second, the $\sigma$-jitter
window has an upper bound at the balance-loss turn-on step
$\rho_b\!\cdot\!N$: whatever per-expert separation has accumulated
by then is pulled back toward uniform by the load-balance penalty,
so $\sigma$ must be aggressive enough for stable specialization to
emerge from shared-basis flattening within $\rho_b\!\cdot\!N$ steps
but not so aggressive that the random initial residual destabilises
early training. This is a fragile two-knob region $(\sigma, \rho_b)$
where neither knob has a principled setting; both are tuned by the
same criterion (router entropy) they are meant to fix. The structural
fix has neither knob: cross-expert orthogonality holds at step~$0$
by construction and survives every gradient step thereafter, so the
balance-loss schedule and the noise scale are both irrelevant to
whether the router de-uniformises.

\paragraph{Comparison with soft specialization regularizers.}
A complementary line of work~\cite{guo2025advancing} attacks the
specialization--uniformity tension with loss-side terms: an
output-orthogonality penalty on selected experts' per-token outputs,
\[
\mathcal{L}_o = \!\!\sum_{i,j\neq k}\!\!
\Bigl\lVert
\frac{\langle \tilde x_{ij}, \tilde x_{ik}\rangle}
     {\langle \tilde x_{ik}, \tilde x_{ik}\rangle + \epsilon}
\,\tilde x_{ik}
\Bigr\rVert^2
\quad\text{with}\quad
\tilde x_{ij} = x_i\,\theta_{E_j}\,\mathbb{1}_{\{s_{ij}>0\}},
\]
and a routing-variance term that rewards decisive gates,
$\mathcal{L}_v = -\sum_{i,j}(s_{ij} - \bar s_j)^2/n$. These coexist
with a standard auxiliary balance loss. The motivating regime is
\emph{pretrained} MoE LLMs in post-training, where experts already
differ non-trivially and the failure mode is regression toward
uniformity under aux-loss pressure.

The cold-start regime that motivates Ortho-Hydra is different.
Experts are freshly initialized ($S_p\!=\!0$, $\lambda\!=\!0$) and
the router sits at the uniform prior, so a per-token
output-orthogonality loss operates on $E$ identical (zero) expert
outputs and a score-variance loss operates on $E$ identical logits.
The gradients of both regularizers are themselves
permutation-symmetric in the expert index $e$ at that initial
condition, and a regularizer cannot break a symmetry its own
gradient shares. Structural cross-expert orthogonality is what the
soft variant approximates asymptotically, made exact and runtime-free
in the parameterization: $\Pbases[i]\T\Pbases[j] = \mathbf{0}$ holds
at step~$0$ by construction and survives every gradient step, with
no activation-memory cost for materialising per-token, per-expert
pre-gate outputs $\tilde x_{ij}$ across the sequence.

The two ideas compose, and the order matters. Of the two soft terms,
only $\mathcal{L}_v$ is well defined at step~$0$: the score-variance
gradient flows into the router weights through the logits regardless
of $\lambda$. $\mathcal{L}_o$ operates on $\tilde x_{ij} = x_i\,
\theta_{E_j}\!\cdot\!\mathbb{1}_{\{s_{ij}>0\}}$, which is identically
zero across experts when $\lambda\!=\!0$, so it contributes no
gradient signal until the residual has lifted off. In Ortho-Hydra
this liftoff happens within the first few hundred steps of
de-uniformisation in Figure~\ref{fig:router-dynamics}. After
liftoff, both terms can ride on top of the disjoint geometry as a
router-side counter-pressure to the late-training balance-loss
pull-back (visible as the small entropy bump at step $11266$ in the
figure, where the balance loss turns on and pulls a specialized
router back toward uniform). A combined experiment is the natural
follow-up: the geometric construction bootstraps the router out of
the deadlock, and $\mathcal{L}_v + \mathcal{L}_o$ defends decisive
routing in the regime where asymptotic output orthogonality is what
the soft variant approximates anyway. The soft fix is logically
posterior to the structural one; output-orthogonality regularization
cannot escape a starting state in which its own direction is
undefined, and that starting state is the default for adapter-MoE on
a frozen backbone.

\paragraph{Timestep-axis specialization has the same shape.}
The reported runs feed the diffusion noise level $\sigma$ into the
router as $4$-bucket sinusoidal features and add a per-bucket
coefficient ($0.3$) to the Switch balance loss, asking each
$\sigma$-bucket's expert distribution to stay near-uniform on its
own, in the hope that different experts would specialize on
different denoising regimes (early high-noise structure vs late
refinement). We did not observe meaningful per-bucket specialization:
per-expert utilisation across $\sigma$-buckets stayed
near-symmetric and the $\sigma$-input columns of the router behaved
as a small additive perturbation on an otherwise content-driven
gate. The diagnosis transfers from the cold-start case. A per-bucket
balance loss is a soft constraint, and it cannot create a
specialization axis the gradient geometry does not already favour:
at the uniform router every $\sigma$-bucket's per-bucket balance
gradient is permutation-symmetric across experts, just as the global
balance gradient is. The structural analogue of disjoint slices is
to partition experts \emph{by} $\sigma$-band at the architecture
level — assign each expert $e$ to a band $e\bmod B$ and mask
out-of-band logits with $-\infty$ before the softmax — so that
\emph{which} experts fire \emph{when} is fixed by construction
rather than by the optimiser. Combined with the disjoint SVD-slice
partitioning, in-band experts remain disjoint in output subspace
while out-of-band experts are disjoint in $\sigma$-coverage. The
implementation ships in the released code with an interleaved
expert-to-band layout, so each band inherits a representative
spread of singular slices instead of binding band $0$ to the top
slice and band $B{-}1$ to the bottom. Band edges are themselves a
design lever: the default uniform $\mathrm{linspace}(0,1,B{+}1)$
weights every $\sigma$-decade equally, but a non-uniform schedule
such as $[0, 0.5, 0.8, 1.0]$ concentrates expert capacity on the
high-$\sigma$ structure-defining regime while sharing a single wide
band across the broad late-refinement window where the velocity
field is approximately linear and per-step adaptation is cheaper.
The dual schedule $[0, 0.2, 0.5, 1.0]$ inverts the choice when
late-step refinement is the bottleneck. Rerunning the cold-start
diagnostic with the hard $\sigma$-partition enabled, and ablating
the band schedule, is left to future work.

\paragraph{Scope of the cold-start experiment.}
Figure~\ref{fig:router-dynamics} tests the mechanism claim at its
strongest resolution: the shared-basis deadlock is directly
observable in router entropy, and the original $\sigma$-jitter
mitigation does not escape it on a matched step budget. The deadlock
is therefore neither a slow-training artefact nor a missing-noise
artefact, but a property of shared-basis gradient geometry that the
structural fix removes. We do not claim end-of-training quality on
multi-style data, sensitivity to $E$, $r$, or the balance-loss
weight, or rank-matched comparisons (the baselines were $r\!=\!32$
versus $r\!=\!48$ for ortho). Whether disjoint slices also improve
end-task quality on multi-style data remains open.

\section{Conclusion}

Ortho-Hydra is a re-parameterization of HydraLoRA in which the
shared down-projection is a Cayley-rotated SVD basis and each expert
owns a disjoint slice of the top-$(Er)$ left singular vectors of the
pretrained weight. Disjointness is preserved exactly through training,
which removes the cold-start symmetry between zero-initialized experts
and gives the layer-local router differentiating gradient signal at
step~$0$. The cold-start router-dynamics experiment
confirms the predicted shared-basis deadlock and shows that the
original $\sigma$-jitter mitigation does not escape it on a matched
budget; end-task quality on multi-style fine-tuning is left to future
work.

\section*{Acknowledgements}

Portions of the text in this manuscript were drafted and edited with
the assistance of large language models (LLMs). The technical
contributions, experimental design, implementation, results, and
conclusions are those of the author, who has reviewed all generated
text and takes full intellectual responsibility for the content of
the paper.

\end{document}